\documentclass[runningheads,a4paper]{llncs}

\usepackage{amssymb}
\usepackage{graphicx}
\usepackage{tabularx}
\usepackage{multibib}
\usepackage{soul}
\usepackage{listings}
\usepackage{color}
\usepackage{dirtytalk}
\usepackage{epstopdf}
\usepackage[latin1]{inputenc}

\usepackage{hyperref}
\usepackage{xstring}

\usepackage{url}
\urldef{\mailsa}\path{rohit.sakala@reseach.iiit.ac.in, navjyoti@iiit.ac.in}    
\newcommand{\keywords}[1]{\par\addvspace\baselineskip
\noindent\keywordname\enspace\ignorespaces#1}

\begin{document}

\mainmatter  % start of an individual contribution

% first the title is needed
\title{Analysis of Speeches in Indian Parliamentary Debates}

% a short form should be given in case it is too long for the running head
\titlerunning{}

% the name(s) of the author(s) follow(s) next
%
% NB: Chinese authors should write their first names(s) in front of
% their surnames. This ensures that the names appear correctly in
% the running heads and the author index.
%
\author{Sakala Venkata Krishna Rohit%
\thanks{}%
\and Navjyoti Singh\\}
\authorrunning{}
% (feature abused for this document to repeat the title also on left hand pages)

% the affiliations are given next; don't give your e-mail address
% unless you accept that it will be published
\institute{Center for Exact Humanities (CEH)\\
International Institute of Information Technology, Hyderabad, India\\
\url{}}

%
% NB: a more complex sample for affiliations and the mapping to the
% corresponding authors can be found in the file "llncs.dem"
% (search for the string "\mainmatter" where a contribution starts).
% "llncs.dem" accompanies the document class "llncs.cls".
%

\toctitle{}
\tocauthor{}
\maketitle

\begin{abstract}
\emph{ With the increasing usage of internet, more and more data is being digitized including parliamentary debates but they are in an unstructured format. There is a need to convert them into structured format for linguistic analysis. Much work has been done on parliamentary data such as Hansard, American congressional floor-debate data on various aspects but less on pragmatics. In this paper, we provide a dataset for synopsis of Indian parliamentary debates and perform stance classification of speeches i.e identifying if the speaker is supporting the bill/issue or is against it. We also analyze the intention of the speeches beyond mere sentences i.e pragmatics in the parliament. Based on thorough manual analysis of the debates, we developed an annotation scheme of 4 mutually exclusive categories to analyze the purpose of the speeches: to find out \textit{ISSUES} , to \textit{BLAME}, to \textit{APPRECIATE} and for \textit{CALL FOR ACTION}.  We have annotated the dataset provided, with these 4 categories and conducted preliminary experiments for automatic detection of the categories. Our automated classification approach gave us promising results. }

\end{abstract}

\section{Introduction}

As the world moves towards increasing forms of digitization, the creation of text corpora has become an important activity for NLP and other fields of research. Parliamentary data is a rich corpus of discourse on a wide array of topics. The Lok Sabha \footnote{Lok Sabha is the lower house of the Indian Parliament.} website \footnote{http://164.100.47.194/Loksabha/Debates/debatelok.aspx} provides access to all kinds of reports, debates, bills related to the proceedings of the house. Similarly, the Rajya Sabha \footnote{Rajya Sabha is the upper house of the Indian Parliament.} website also contains debates, bills, reports introduced in the house. The Lok Sabha website also contains information about members of the parliament who are elected by the people and debate in the house. Since the data is unstructured , it cannot be computationally analyzed. There is a need to shape the data into a structured format for analysis. This data is important as it can be used to visualize person, party and agenda level semantics in the house. \\

The data that we get from parliamentary proceedings has presence of sarcasm, interjections and allegations which makes it difficult to apply standard NLP techniques \cite{Onyimadu-14}. Members of the parliament discuss various important aspects and there is a strong purpose behind every speech. We wanted to analyze this particular aspect. Traditional polar stances (for or against) do not justify for the diplomatic intricacies in the speeches. We created this taxonomy to better understand the semantics i.e the pragmatics of the speeches and to give enriched insights into member's responses in a speech. The study of the speaker's meaning, not focusing on the phonetic or grammatical form of an utterance, but instead on what the speaker's intentions and beliefs are is pragmatics. Pragmatics is a sub-field of linguistics and semiotics that studies the ways in which context contributes to meaning.\\

After thorough investigation of many speeches we found that the statements made by members cannot be deemed strictly "for or against" a bill or government. A person maybe appreciating a bill or government's effort in one part of a speech but also asking attention to other contentious issues. Similarly, a person criticizing government for an irresponsible action could be giving some constructive suggestions elsewhere. A political discourse may not always be polar and might have a higher spectrum of meaning. After investigating and highlighting statements with different intentions we came up with a minimal set of 4 mutually exclusive categories with different degrees of correlation with the traditional two polar categories (for and against). It is observed that any statement by a participating member will fall into one of these categories namely - Appreciation, Call for Action, Issue, Blaming.\\

\begin{table}[h!]
\begin{center}
\begin{tabularx}{\columnwidth}{|l|X|X|X|X|}

      \hline
      \textbf{Stance/Category} & \textbf{Appreciation}  & \textbf{CallForAction}&  \textbf{Issue} & \textbf{Blame}  \\
      \hline
      For & Medium & High & Low  & Low \\
      \hline
      Against & Low & High & High & Low \\ 
      \hline
\end{tabularx}
\caption{Degrees of Co-Relation}
\end{center}
\end{table}

For example, if the debate consists of more of issues, one can infer that the bill is not serving the its purpose in a well manner. Also, this preliminary step will lead to new areas of research such as detection of appreciation, blame in similar lines of argument mining which is evolving in the recent years in the field of linguistics. We will quote portions of a few speeches which will give an idea of the data being presented: \\

\say{\textit{This city has lost its place due to negligence of previous governments and almost all industries have migrated from here and lack of infrastructure facilities, business is also losing its grip. It is very unfortunate that previous UP Governments also did not do any justice to this city.  }}  \\ 

\hspace{12ex}  - Shri Devendra Singh Bhole, May 03, 2016  \\

As evident, the speaker is clearly blaming the previous governments for negligence on the city. In this sense the data is very rich and a lot of linguistic research is possible. Researchers can work on different aspects such as detection of critique made by members, suggestions raised by members etc. Given the data, it can be used for rhetoric, linguistic, historical, political and sociological research. Parliamentary data is a major source of socially relevant content. A new series of workshops are being conducted for the sole purpose of encouraging research in parliamentary debates \href{https://www.clarin.eu/ParlaCLARIN}{\color{blue}{ParlClarin}}. \\

As a preliminary step, we created four major categories of the speeches spoken by the parliament members. The definitions and examples of the four categories are explained in the below tables respectively. The examples are taken from a debate on NABARD bill in Lok Sabha. \\

\begin{table}[h!]
\begin{center}
\begin{tabularx}{\columnwidth}{|l|X|}

      \hline
      \textbf{Category} & \textbf{Definition} \\
      \hline
      Issue & raise problems in general which need attention. \\
      \hline
      Blame & blaming the government or the opposition government or policies.  \\
      \hline
      Appreciate & appreciating and justifying policies , governments by mentioning the benefits and efforts of it. \\
      \hline
      Call for Action & speeches in which members suggest, request for new laws/proposals. \\
      \hline

\end{tabularx}
\caption{Definition of the categories}
\end{center}
\end{table}

\begin{table}[!h]
\begin{center}
\begin{tabularx}{\columnwidth}{|l|X|}
      
      \hline
      \textbf{Category} & \textbf{Example Statements} \\
      \hline
      Issue & ....The present scenario is that credit from nationalized banks and from financial institutions is not available. That is why there is a crisis in the agrarian economy.....\\
      \hline
      Blame & .....The policy of the Government is going in one direction and the banks which come under the Finance Ministry are going in a  totally different directions.....\\
      \hline
      Appreciate & .....The Government's vision of 'Sabka Sath Sabka Vikas' comes to the fore through this Bill......This will usher in not only the rural development but also leave the rural folk to their self-reliance. At the same time,  the  rural  development  will  prevent  them  from migrating to the cities and everybody will get employment..... \\
      \hline
      Call for Action & .....So, NABARD should come forward with new credit and interest policy so that more farmers could avail of the loan.....That is why, the coverage of NABARD and its access to the real masses or the poor people should be further expanded.\\
      \hline

\end{tabularx}
\caption{Example statements of the categories}
\end{center}
\end{table}

A speech can be labelled with multiple categories as members can appreciate and raise issues in the same speech. The following points are the contributions of this paper :

\begin{itemize}
  \item A dataset of Indian Parliamentary Debates.
  \item Creation of categories for classifying the purpose of the speech acts i.e pragmatics
  \item Preliminary experiment on automatic detection of the categories.
  \item Stance classification of speeches. 
\end{itemize}

\section{Related Work}

Many linguists around the globe are concentrating on creation of parliamentary datasets. \cite{Darja-2017} gives an overview of the parliamentary records and corpora from countries with a focus on their availability through Clarin infrastructure. A dataset of Japanese Local Assembly minutes was created and analyzed for statistical data such as number of speakers, characters and words \cite{kimura-16}. \cite{Najeh-14} created a highly multilingual parallel corpus of European parliament and demonstrated that it is useful for statistical machine translation. Parliamentary debates are full of arguments. Ruling party members refute the claims made by opposition party members and vice versa. Members provide strong arguments for supporting their claim or refuting other's claim. Analyzing argumentation from a computational linguistics point of view has led very recently to a new field called argumentation mining \cite{green_14}. One can perform argument mining on these debates and analyze the results. \cite{David-17} worked on detecting perspectives in UK political debates using a Bayesian modelling approach. \cite{Lippi-2016} worked on claim detection from UK political debates using both linguistic features text and features from speech.

Stance classification is a relatively new and challenging approach to deepen opinion mining by classifying a user's stance in a debate i.e whether he is for or against the topic. \cite{addawood-2017}.  \cite{Thomas-06} addressed the question of whether opinion mining techniques can be used on Congressional debates or not. \cite{Getoor-2014} worked on stance classification of posts in online debate forums using both structural and linguistic features. \cite{Kerren-2016} trained a svm \cite{Pedregosa-11} classifier with features of unigrams, bigrams and trigrams to predict whether a sentence is in agreement or disagreement and achieved an F-score of 0.55 for agreement and 0.81 for disagreement on the evaluation set. No one has worked on classifying speeches based on their purpose. This is the first novel work towards this aspect.

\section{DataSet}

Our dataset consists of synopsis of debates in the lower house of the Indian Parliament (Lok Sabha). The dataset consists of :

\begin{itemize}
  \item 19 MB approx.
  \item 189 sessions
  \item 768 debates
  \item 5575 speeches
  \item 1586838 words
  \item Sessions from 2014 to 2017.
\end{itemize}

In Lok Sabha, a session is referred to as all the debates held in a particular cycle of sitting. There are 55 debate types \footnote{http://164.100.47.194/Loksabha/Debates/DebateAdvSearch16.aspx} identified by the Lok Sabha. Table 3 identifies some of the debate types we have considered and their frequency between the years 2014 and 2017. We opted out debate types which do not occur regularly. Each debate type has its own style of proceedings. For example, in the debate type "Government Bills", a minister places a bill on the table and discussion is carried out on the bill where as in the debate type "Matter under 377", each speaker raises an issue of which he is concerned of but no discussion is done on the issues.

\begin{table}[!h]
\begin{center}
\begin{tabularx}{\columnwidth}{|l|X|}

      \hline
      \textbf{Debate Type} & \textbf{Occurrence Count}\\
      \hline
      Matter Under 377 & 124 \\
      \hline
      Submission Members & 101\\
      \hline
     Government Bills & 87\\
     \hline
     Discussion & 42\\
      \hline
      General & 41 \\
     \hline
      References & 37 \\
      \hline
      Minister Statement & 33 \\
      \hline
     Statutory Resolutions & 27 \\
     \hline
     Private Member Bills & 15 \\
     \hline
     President Thanks & 12 \\
     \hline
     Motion & 6 \\
     \hline
    
\end{tabularx}
\caption{Different debate types in Lok Sabha.}
 \end{center}
\end{table}

\subsection{Creation}

The creation of the dataset involved 3 steps. The first step was to scrap the pdf files from the Lok Sabha website. Each pdf file is a session. The second step was to convert the pdf files into text files for easy parsing. The challenge here was to convert this unstructured information into a structured format. The third step is to extract relevant data using pattern matching. We developed a software parser \footnote{https://github.com/rohitsakala/synopsisDebateParser} for extracting the entities such as date, debate type, member name and speech. We used regex, pattern matching code to find out patterns from the text file. For example to segregate a speaker's name from his speech, we used :

\medskip

\noindent
{}
\begin{verbatim}
re.split(":")
\end{verbatim}
\noindent
{\small}

as name of the speaker and his/her speech is separated by a colon. An example pdf can be accessed using this \href{http://164.100.47.193/Synop/16/XIV/Sup+Synopsis-08-02-2018.pdf}{\color{blue}{URL}} . Right now, member name and bill name are needed to be stored manually which we plan to automate too. Sometimes the pattern matching fails due to irregularities in the pdf as those were written by humans though they were negligible. We stored the structured data into a Mongo database as different debate types have different schema. The database consists of the following tables :

\begin{itemize}
    \item{ \texttt{Sessions} : all the debates happened on a particular day with date, secretary general name. }
    \item{ \texttt{Members}  : information about the members/speakers of the parliament i.e name and party affiliation. }
    \item{ \texttt{Debates}  : contains the member id and the corresponding speeches, summaries and keywords. }
    \item{ \texttt{Bills}    : the name of the bill. }
    \item{ \texttt{Debate Type}    : the name of the debate type. }
\end{itemize}

The software parser developed is very generic. As new sessions are being added on the Lok Sabha website, the software parser automatically identifies them, parses it and stores the structured data in the database. The database has been hosted in a online database hosting site,  \href{https://mlab.com}{\color{blue}{mLab}}. The mongo shell can be accessed using this command in any linux machine which has mongo installed.

\medskip

\noindent
{}
\begin{verbatim}
mongo ds235388.mlab.com:35388/synopsis -u public -p public
\end{verbatim}
\noindent
{\small}

\subsection{Annotation}

We have annotated 1201 speeches with the four categories mentioned above, on the speeches. We also annotated stances of the speakers towards the bill/issue that is being debated on. There are two stances one is \textit{for} and other is \textit{against}. The statistics of the annotated data is shown in Table 4.

\begin{table}[!h]
\begin{center}
\begin{tabularx}{\columnwidth}{|l|X|}

     \hline
     \textbf{Categories} & \textbf{Count}\\
     \hline
     Issue & 589\\
     \hline
     Blame & 147\\
     \hline
     Appreciate & 522\\
     \hline
     Call for Action & 930\\
     \hline
     For & 919\\
     \hline
     Against & 282\\
     \hline
    
\end{tabularx}
\caption{Statistics of Annotated Data}
 \end{center}
\end{table}

Two humanities students were involved in the annotation of the four categories on 1201 speeches. The annotator agreement is shown in Table 5 and is evaluated using two metrics, one is the Kohen's Kappa \cite{Jacob-1960} and other is the inter annotator agreement which is the percentage of overlapping choices between the annotators.

\begin{table}[!h]
\begin{center}
\begin{tabularx}{\columnwidth}{|l|X|X|}

     \hline
     \textbf{Category} & \textbf{Kohen's Kappa} & \textbf{Inter Annotator Agreement} \\
     \hline
     Issue & 0.67 & 0.84 \\
     \hline
     Blame & 0.65 & 0.90 \\
     \hline
     Appreciate & 0.88 & 0.94 \\
     \hline
     Call for Action & 0.46 & 0.92 \\
     \hline
    
\end{tabularx}
\caption{Inter Annotator agreement metrics of Annotated Data}
 \end{center}
\end{table}

The inter annotator agreement for the stance categories were 0.92. The high values of inter annotator scores clearly explain how easy it was to delineate each category. It also signifies that the definition of the category that needed to be annotated, were very clear.

\subsection{Keywords and Summarization}

We have used TextRank which is an extractive summariser \cite{Mihalcea-04} for summarizing the entire debate and for finding keywords in the debate. TextRank is a graph based ranking model for text processing specifically KeyPhrase Extraction and Sentence Extraction. TextRank performs better in text summarization using graph based techniques \cite{Mihalcea-041}. We added these two extra fields i.e the keywords extracted by TextRank and the summary created by TextRank in the debates collection. An example summary is : \\

\textit{The last National Health Policy was framed in 2002.
The Policy informs and prioritizes the role of the Government in shaping health systems in all its dimensions investment in health, organization and financing of health care services, prevention of diseases and promotion of good health through cross-sectoral action, access to technologies, developing human resources, encouraging medical pluralism, building the knowledge base required for better health, financial protection strategies and regulation and progressive assurance for health.
The Policy aims for attainment of the highest possible level of health and well-being for all at all ages, through a preventive and promotive health care orientation in all developmental policies, and universal access to good quality health care services without anyone having to face financial hardship as a consequence.
The Policy seeks to move away from Sick-Care to Wellness, with thrust on prevention and health care promotion.
Before this, the Policy was for the Sick-Care Health Policy.
Now we are making it Promotional and Preventive Health Policy.
While the policy seeks to reorient and strengthen the public health systems, it also looks afresh at strategic purchasing from the private sector and leveraging their strengths to achieve national health goals.
As a crucial component, the policy proposes raising public health expenditure to 2.5 per cent of the GDP in a time bound manner.
The Policy has also assigned specific quantitative targets aimed at reduction of disease prevalence/incidence under three broad components viz., (a) health status and programme impact, (b) health system performance, and (c) health systems strengthening, aligned to the policy objectives.
To improve and strengthen the regulatory environment, the policy seeks putting in place systems for setting standards and ensuring quality of health care.
The policy advocates development of cadre of mid-level service providers, nurse practitioners, public health cadre to improve availability of appropriate health human resource.
The policy also seeks to address health security and Make in India for drugs and devices.
It also seeks to align other policies for medical devices and equipment with public health goals.}

\subsection{Detection of Polarity}

To detect the polarity of each speech, we have used VADER \cite{Hutto-2014} sentiment analysis tool. The tool uses a simple rule-based model for general sentiment analysis and generalizes more favorably across contexts than any of many benchmarks such as LIWC and SentiWordNet. The tool takes as input a sentence and gives a score between -1 and 1. The polarity of a speech is calculated by taking the sum of the polarities of the sentences. If the sum is greater than zero, then it is classified as \textit{positive}, if it is less than zero, then it is classified as \textit{negative} and if it is equal to zero then it is classified as \textit{neutral}. The statistics of the data is presented in Table 6. 

\begin{table}[!h]
\begin{center}
\begin{tabularx}{\columnwidth}{|X|X|}

     \hline
     \textbf{Category} & \textbf{Count}\\
     \hline
     Positive & 4006 \\
     \hline
     Negative & 1457 \\
     \hline
     Neutral & 112 \\
     \hline
     Total & 5575 \\
     \hline
    
\end{tabularx}
\caption{Sentiment Polarity of Speeches}
 \end{center}
\end{table}

\subsection{Examples}

\begin{itemize}

\item{
A Document in session collection\footnote{A table in mongo database is called collection.}.
}

\end{itemize}

\begin{lstlisting}[basicstyle=\footnotesize]
{
"_id" : ObjectId("5a4255c789.."),         
"indianDate" : "Vaisakha 9,1938(Saka)",
"debates" : {
"5999649837.." : ObjectId("5a425b5.."),
"5999644a37.." : ObjectId("5a425b06..")
}
"englishDate" : "Friday,April 29,2016",
"houseName" : "LOK SABHA",
"secretaryGeneralName" : "ANOOP MISHRA"
}
\end{lstlisting}

The \_id is the unique key assigned by the mongo database. The keys\footnote{A key refers to the key in the json data type.} in the debates key represent the debate types from the debate types collection. The values of the debates key refer to the corresponding debates in the debates collection. 

\begin{itemize}

\item{
A Document in member collection. The table consists of name of the member spoken, the house of the parliament and the party to which he is affliated. 
}

\end{itemize}

\begin{lstlisting}
{
"_id" : ObjectId("59a8e0e983"),
"name" : "Dharambir Singh,Shri",
"house" : "Lok Sabha",
"party" : "BJP" 
}
\end{lstlisting}

\begin{itemize}

\item{
A Document in bill collection. The table consists of the bill name.
}

\end{itemize}

\begin{lstlisting}
{
"_id" : ObjectId("59de525596..."),
"name" : "THE COMPENATION BILL, 2016"
}
\end{lstlisting}

\begin{itemize}

\item{
A Document in debates collection of debate type Submission Members. The table consists of all the speeches made in a particular debate in an order with summary and keywords from TextRank.
}

\end{itemize}

\begin{lstlisting}
{
"_id" : ObjectId("5a42539889.."),
"topic" : "Flood situation in ...",
"keywords" : "water state ... ",
"summary" : "...",
"speeches" : {
  "1" : {
  "speech" : "In Tamil Nadu and in...",
  "memberId" : "59a92d88a0b4...",
  "polarity" : "Negative"
  },
  "2" : {
  "speech" : "We all have witness...",
  "memberId" : "59cbc3ef6636...",
  "polarity" : "Positive"
  },
  "3" : {
	...
  }
  ...
  ...
}

\end{lstlisting}

The memberId refers to the \_id in the member's collection.
\section{Experiment}

In this section, we deal with two tasks,  task one is the classification of the stances the speakers take and task two is the classification of categories based on purpose. Stance classification differs from sentiment analysis. For instance, the number of speeches that were annotated as \textit{for} i.e 919 had only 719 labelled as \textit{positive} and the number of speeches that were annotated as \textit{against} i.e 282 had only 89 as \textit{negatively} labelled. So, these statistics clearly indicate the difference between polarity detection and stance classification.\\

Text classification is a core task to many applications, like spam detection, sentiment analysis or smart replies. We used fastText and SVM \cite{Hearst-1998} for preliminary experiments. We have pre-processed the text removing punctuation's and lowering the case. Facebook developers have developed fastText \cite{joulin-2016} which is a library for efficient learning of word representations and sentence classification. The reason we have used fastText is because of its promising results in \cite{Jha-17}. \\

We divided our training and testing data in the ratio of 8:2 for classification. As mentioned above we used fastText and SVM for both the classification tasks. We report accuracy for each class as it is a multi-label classification problem. The results are shown in Table 7 and Table 8. Also, the parameters used for fastText is described in Table 9.

\begin{table}[!h]
\begin{center}
\begin{tabularx}{\columnwidth}{|X|X|X|}
      \hline
      \textbf{Task/Metric} & \textbf{fastText} & \textbf{SVM} \\
      \hline
      For/Against & \textbf{0.80} & 0.76 \\
      \hline

\end{tabularx}
\caption{Accuracy Score for Classification Task 1}
\end{center}
\end{table}

\begin{table}[!h]
\begin{center}
\begin{tabularx}{\columnwidth}{|l|X|X|r|}
      \hline
      \textbf{Task 2/Metric} & \textbf{fastText} & \textbf{SVM} \\
      \hline
      Call for Action & \textbf{0.745} & 0.72 \\
      \hline
      Issue &  \textbf{0.604} & 0.56 \\
      \hline
      Blame & 0.783 & \textbf{0.84} \\
      \hline
      Appreciate & \textbf{0.679} & 0.62 \\
      \hline

\end{tabularx}
\caption{Accuracy Score for Classification Task 2}
\end{center}
\end{table}

\begin{table}[!h]
\begin{center}
\begin{tabularx}{\columnwidth}{|l|X|}
      \hline
      \textbf{Parameter} & \textbf{Value} \\
      \hline
      Learning Rate & 0.8\\
      \hline
      Word Dimension & 100\\
      \hline
      n-gram & 2\\
      \hline
      Epoch & 100\\
      \hline
      Loss Function & hs\\
      \hline

\end{tabularx}
\caption{Parameters used for fastText algorithm}
\end{center}
\end{table}

We have not used hs (Hierarchical Soft-max) for binary classification, instead used regular softmax as it was giving better results in fastText. \\

For SVM, the features were the word vectors trained using word2vec \cite{Mikolov-2013} with dimesion size of 300 whereas for fastText, the features were the word vectors trained using character n-gram embedding. We have achieved considerably good results. We plan to annotate more and check if the accuracy increase any further. The limitation that we feel is the number of annotations being done. We approached the classification problem as one vs rest classification problem. We performed the classification on document level. Later we would like to analyze at sentence level. The least accuracy was for Issue category and the highest is for Blame category. This research will inspire researchers to take on further research on mining appreciation, blaming from text in lines with the ongoing approaches of argument mining, hate speech, sarcasm generation etc. \\

As we increase the number of epochs in the fastText, the scores also increase as evident from Table 10, but the increase stops after 25 epochs. 

\begin{table}[!h]
\begin{center}
\begin{tabularx}{\columnwidth}{|X|X|}
      \hline
      \textbf{Epochs} & \textbf{Accuracy} \\
      \hline
      5 & 0.579 \\
      \hline
      10 & 0.65\\
      \hline
      25 & 0.6916 \\
      \hline
      50 & 0.6708 \\
      \hline
      100 & 0.679 \\
      \hline

\end{tabularx}
\caption{Analysis of fastText for Task 2 with varying epochs for Call for Appreciate category}
\end{center}
\end{table}

\section{Conclusion}

In this paper, we presented a dataset of synopsis of Indian parliamentary debates. We developed a generic software parser for the conversion of unstructured pdfs into structured format i.e into a relational database using mongo database software. We analyzed the purpose of the speeches of the member of parliament and categorized them into 4 major categories and provided statistics of the categories. We also tried to identify them automatically using fastText algorithm and provided the results. The analysis is done for understanding the purpose of the speeches in the parliament. We also presented our results on binary stance classification of the speeches whether the member is in favour of the debate topic or not.

\section{Future Work}

In future, we would like to increase the size of the dataset by including sessions of previous years which are not yet digitized. Sessions before 2009 are yet to be digitalised by the Lok Sabha editorial of India. Also we plan to include Rajya Sabha debates into the dataset. We have used fastText for classifying categories. We plan to develop a set of features to increase the accuracy of the classification task as we believe that features like party affiliation will have greater impact and experiment with other machine learning approaches. \\

TextRank is used for summarization. We feel that for political debates, summarization should emphasize on arguments made by members unlike TextRank. In the whole debate, a lot of themes are raised by the members. The debate revolves around these themes. So, developing a model for thematic summarization with arguments will capture the complete picture of the entire debate unlike TextRank. We plan to do this as our future work on these debates. A short summary of the important themes discussed with its arguments will benefit journalists, newspaper editors, common people etc.

\end{document}